\documentclass[runningheads]{llncs}
\usepackage[T1]{fontenc}
\usepackage{graphicx}
\usepackage{subcaption}
\usepackage{amsmath, amssymb}
\usepackage{multirow}
\usepackage{tikz}
\usepackage[bottom]{footmisc}

\definecolor{tabblue}{HTML}{1f77b4}
\definecolor{taborange}{HTML}{ff7f0e}
\definecolor{tabgreen}{HTML}{2ca02c}
\definecolor{tabred}{HTML}{d62728}
\definecolor{tabpurple}{HTML}{9467bd}
\definecolor{tabbrown}{HTML}{8c564b}

\newcommand*{\tikzblock}[1]{%
  \begin{tikzpicture}
    \draw[draw=#1, fill=#1] (-.3em,-.3em) rectangle (.3em,.3em);
  \end{tikzpicture}%
}
\begin{document}
\title{A New Angle on Bones: Robust Pose Estimation in X-Ray and Ultrasound}
\titlerunning{Robust Inter-Bone Pose Estimation in X-Ray and Ultrasound}

\author{Ron Keuth\inst{1}\orcidID{0009-0003-4289-2836} \and
Christoph Großbröhmer\inst{1,4}\orcidID{0000-0002-8926-8729} \and
Franziska Halm\inst{2} \and
Miriam Johann\inst{3} \and
Anne-Nele Schröder\inst{3} \and
Ludger Tüshaus\inst{3}\orcidID{0000-0001-7997-6396} \and
Mattias P. Heinrich\inst{1}\orcidID{0000-0002-7489-1972} \and
Lasse Hansen\inst{4}\orcidID{0000-0003-3963-7052}}

\authorrunning{R. Keuth et al.}


\institute{Medical Informatics, University of Lübeck, Ratzeburger Allee 160,
23562 Lübeck, Germany. \and
Institut of Radiology and Nuclear Medicine, University Hospital Schleswig-Holstein \and
Paediatric Surgery, University Hospital Schleswig-Holstein \and
EchoScout GmbH, Maria-Goeppert-Str. 3, 23562 Lübeck, Germany
}

\maketitle              
\begin{abstract} 
Measuring the angle between bone structures is a routine task in medical image analysis and provides a key quantitative parameter for diagnosis and treatment planning. Automated methods can reduce time and cost while improving reproducibility.
In this work, we address automatic bone pose estimation using a learning-based point candidate proposal followed by a line model to extract axis parameters.
Since conventional line models such as least squares are sensitive to outliers, we incorporate false-positive reduction strategies and robust fitting techniques, such as  RANSAC and Hough transforms, to improve robustness.
We evaluate our method on three  clinically relevant paediatric angle estimation tasks: fracture fragment assessment in radiographs and ultrasound and developmental dysplasia of the hip evaluation in ultrasound using the Graf method.
Our approach achieves mean errors of $4.1^\circ$, $5.4^\circ$, and $5.51^\circ$, respectively, not only remaining within the expected clinical observer variability, but also significantly outperforming landmark-based methods.
Our code and annotations for fracture angle assessment in radiographs are publicly available on GitHub.
\end{abstract}
\section{Introduction}
In medical image analysis, measuring the angle between two anatomical structures is a routine task, as it provides a key quantitative parameter for diagnosis, treatment planning, and monitoring the progression of treatment or the healing process over time.
Bones are particularly suitable anatomical structures for this purpose, as they are rigid and clearly visible in medical imaging like radiographs or ultrasound.
Especially in the paediatric domain, bone pose estimation provides a diagnostic foundation, where it is used to monitor the skeleton development of the child.
A prominent example is the sonographic screening for developmental dysplasia of the hip (DDH) through Graf's technique in infants as a routine examination performed at the age of four weeks in Germany (U3).
DDH is the most common skeletal anomaly globally, with an incidence of approximately $2-4~\%$, making early diagnosis critical to avoid corrective surgery in adulthood.
The Graf method quantifies hip joint morphology via two angles measured in a standardized coronal ultrasound plane: the alpha angle, reflecting the bony acetabular roof inclination, and the beta angle, characterizing the cartilaginous roof, which together enable classification of hip maturity and dysplasia severity \cite{kolovos2024grafMethod}.
Angle measurement is also a key parameter in paediatric fracture treatment planning, where it quantifies the angle between fragments.
Distal forearm fractures are the most common injuries in children and adolescents, with approximately $800\,000$ cases treated annually in Germany \cite{Krause2005}.
Due to the high remodelling potential of the growing skeleton, treatment strategies differ fundamentally from those in adults.
While adult fractures typically require surgical realignment, many paediatric fractures can be managed conservatively with immobilization in a plaster cast.
With this, the risks of general anaesthesia can be avoided, while also reducing psychoemotional stress for the child and family, as well as overall healthcare costs.
While radiography is the standard image modality to diagnose paediatric wrist fractures \cite{nagy_pediatric_2022}, ultrasound (US) has emerged as a fast, versatile, and cost-effective alternative involving no ionizing radiation.
Its application to the distal forearm anatomy is well-supported, with clinical studies demonstrating diagnostic performance on par with conventional radiography \cite{ackermann2019sokrat}, resulting in US being the recommended diagnostic tool for suspected wrist fractures in children \cite{AWMF2023}.\par
However, due to US-specific noise and artefacts, robust quantitative angulation assessment of fracture fragments and in the context of DDH remains challenging.
Furthermore, since ultrasound only visualizes the cortical bone surface rather than the full bone cross-section, established angle measurement methods from radiography do not directly transfer to ultrasound images.
Besides those prominent paediatric applications, bone pose estimation is employed to quantify limb alignment via the hip-knee-ankle angle \cite{insall1979knee}, classifying varus and valgus knees, or diagnosing foot deformities via the hallux valgus angle or the calcaneal pitch (distinguishing flatfoot vs high arch) \cite{liszka2018hallux}.
Other assessments include different inter-structure hip geometries like the neck-shaft angle and angle between femoral head and neck axis \cite{Pearle2009LowerLimb}, reflecting the overall bone alignment and biomechanics to indicate risks for unbalanced load distribution and early osteoarthritis.
The Cobb angle \cite{cobb1948outline} represents the spine curvature to diagnose scoliosis and is also an important factor when monitoring its treatment.\par

Considering the broad range of clinical use cases, automated angle estimation can significantly reduce clinician workload.
Manual measurements, by contrast, are subjective and prone to intra- and inter-observer variability, with errors of up to 5° and differences reaching 8° \cite{langensiepen2013scoliosissurvey,gstoettner2007inter}.
Automated methods therefore are not only time and cost-efficient, but also provide more reproducible and consistent results independent of individual interpretation.

\subsection{Related Work}
Automatic bone pose estimation is an active area of research, with most existing work focusing on angle estimation in clinical contexts of developmental dysplasia of the hip (DDH) and the Cobb angle for scoliosis.
For the latter, the Accurate Automated Spinal Curvature Estimation MICCAI Challenge (AASCE) \cite{Wang2021EvaluationAC} provides an official benchmark for comparison of newly proposed methods.
In general, existing approaches can be broadly categorized into two main groups:
The first focuses on localizing landmarks, which are subsequently used to construct axes and compute the angle of interest.
The second group comprises segmentation-based methods, where the bone axis is either predicted directly or derived from the segmentation via post-processing, typically using linear regression to estimate line parameters.

\subsubsection{Landmark-based Approaches}
To implement landmark regression, several approaches for DDH \cite{Li2022DeepLA,Chen2023AutomaticAH} and Cobb angle estimation \cite{Caesarendra2022AutomatedCA} adopt heatmap regression \cite{bulat_heatmap_regression_2016}.
For Cobb angle estimation, landmark-based methods have been further extended to improve robustness.
This includes techniques for radiograph denoising \cite{Zhang2019AnAC}, modelling interdependencies between vertebrae \cite{Zhang2021MPFnetAE}, and fusing angle predictions from anteroposterior and lateral radiographs in a learning-based manner \cite{Wang2019AccurateAC}.
In contrast to heatmap-based approaches, some methods directly regress landmark coordinates and have shown promising performance for Cobb angle estimation \cite{Lin2022Seg4RegCL} as well as for the critical shoulder angle \cite{Minelli2022MeasuringTC}.
Other works enhance landmark detection by incorporating additional representations such as segmentation maps \cite{Fu2020AnAE,Hu2021JointLA}, or by integrating classification tasks \cite{Xu2022ADA}.

\subsubsection{Segmentation-based Approaches}
In the context of DDH, landmarks defining reference axes are extracted from segmentation maps of the ilium and labrum using classical image-processing techniques such as local peak detection and skeletonization \cite{Tasolar2025ADL}.
For Cobb angle estimation, spinal segmentations are used to fit tangent lines along the vertebral column, enabling curvature measurement \cite{Tu2019AutomaticMA}.
Similarly, in hallux valgus classification, the toe axis is derived from segmentation using contour-based features and local peak detection followed by least-squares regression to estimate axis parameters \cite{Kwolek2019MeasuringTA}.
In contrast, \cite{Takeda2024AutomaticEO} avoids manual feature engineering by directly predicting the axis as a thin segmentation line, with subsequent regression for parameter estimation.
While the last two approaches are conceptually similar to ours, they lack robustness mechanisms and thus being sensitive against outliers.
Similarly, approaches based on classical feature extraction like the Canny edge detector require outlier-robust line fitting methods, e.g., the Hough transform, to reliably detect bone axes in DDH \cite{AlBashir2015AlgorithmFA}.
Rather than explicitly estimating axes in the context of Cobb angle assessment, alternative approaches first localize individual vertebrae and subsequently infer their orientation by sampling from a densely predicted vector field that encodes local tilt directions \cite{Zou2023VLTENetAD,Kim2020AutomationOS}.\par
A recent approach targeting ultrasound images of paediatric distal forearm fractures combines segmented bone boundaries, landmarks derived from the segmentation, and an optimization-based alignment of paired dorsal and palmar views enables angulation measurement in the sagittal plane \cite{liu2025automatic}.
However, the method is limited to this single plane, leaving angulation in the frontal plane unaddressed.

\subsection{Contribution}
In this work, we present a method for automatic bone pose estimation that represents bone axes as dense point candidates predicted by a deep learning model, enabling more robust fitting compared to two-point landmark regression. 
We systematically compare three line-fitting models (PCA, RANSAC, Hough transform) and two false-positive reduction strategies with dataset-specific hyperparameter optimization for a fair comparison.
We evaluate our method on three clinically relevant paediatric tasks in two imaging modalities, demonstrating consistent improvements over the landmark baseline. 
In addition to our code, we release new oriented bounding box annotations for fracture fragment angle estimation on the public GRAZPEDWRI-DX dataset\footnote{\url{github.com/multimodallearning/RobustBonePoseEstimation}\label{fn:githublink}}.

\section{Method}
\begin{figure}[b!]
    \centering
    \newcommand{\imgscale}{0.95}
    \begin{subfigure}[t]{.3\linewidth}
        \includegraphics[width=\imgscale\linewidth]{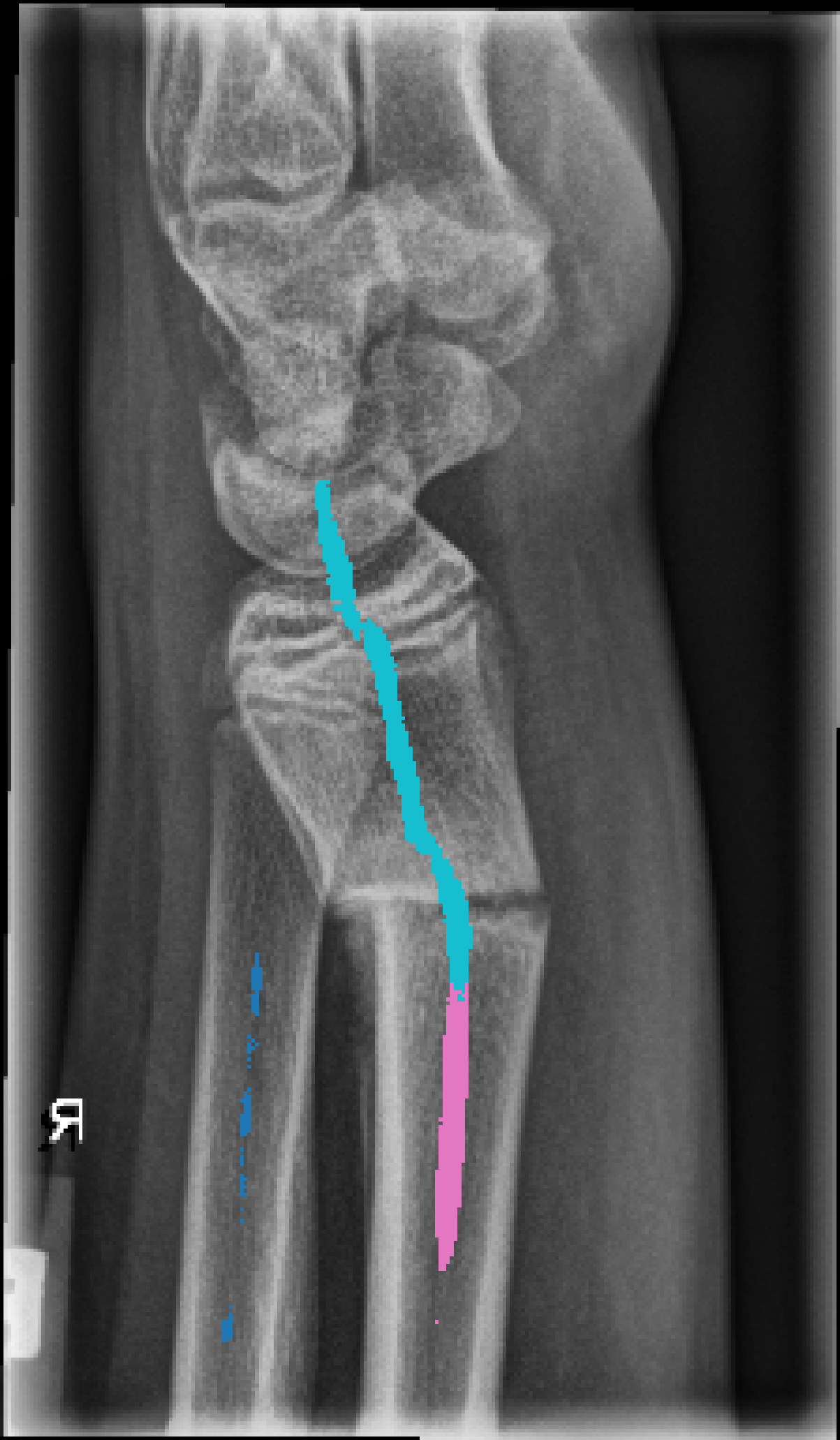}
        \caption{}
    \end{subfigure}%
    \hfill
    \begin{subfigure}[t]{.3\linewidth}
        \includegraphics[width=\imgscale\linewidth]{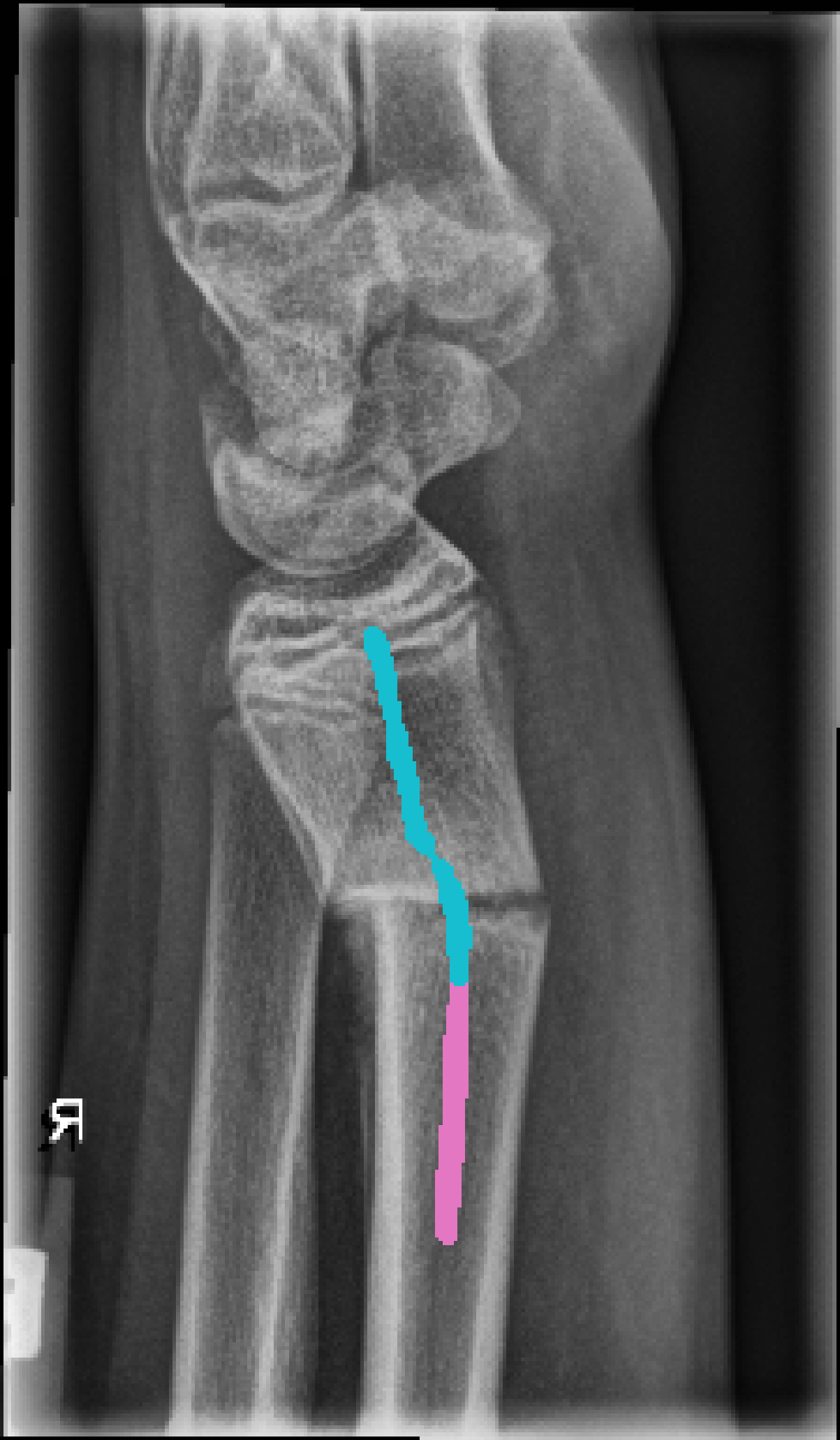}
        \caption{[optional]}
    \end{subfigure}%
    \hfill
    \begin{subfigure}[t]{.3\linewidth}
        \includegraphics[width=\imgscale\linewidth]{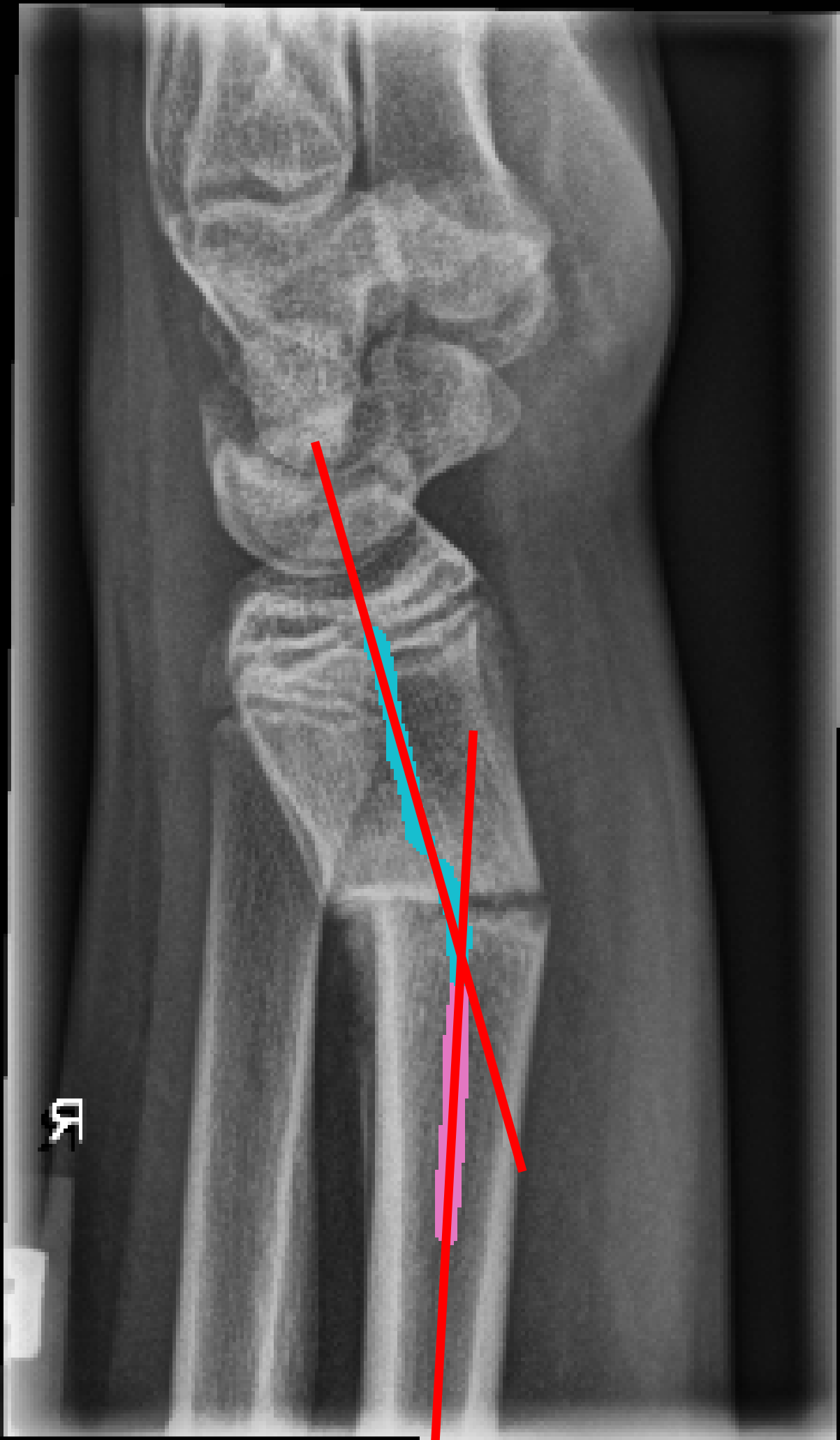}
        \caption{}
    \end{subfigure}
    \caption{Overview of proposed method for bone pose estimation with fracture fragments as example: (a) Predict point candidates with deep learning, (b) False-positive reduction (FPR) (only applied for PCA model), (c) Robust axis estimation via line models like FPR+PCA, Hough transformation, or RANSAC.}
    \label{fig:method_overview}
\end{figure}

Fig.~\ref{fig:method_overview} provides an overview of our proposed method for bone pose estimation.
A deep learning model proposed point candidates for each bone structure of interest.
From those candidates, a robust line model extracts the axes, allowing us to compute their angle.
Depending on the line model, we also introduce different commonly used methods for false-positive reduction.

\subsection{Formulation of Angle Estimation}\label{sec:line_definition}
To measure the angle between two bones, their axes are approximated by fitting lines and computing the angle between them.
A line $\mathcal{L}$ is defined by a point $\mathbf{p}_0 = [x_0, y_0]^\top$ and a unit direction vector $\hat{\mathbf{v}} = [x_v, y_v]^\top$, such that $\mathcal{L} = \{\mathbf{p}_0 + \lambda \hat{\mathbf{v}} \mid \lambda \in \mathbb{R}\}$.
Given two distinct points $\mathbf{p}_1 \neq \mathbf{p}_2$, the parameters can be obtained by
\begin{equation}\label{eq:two_pnts_line}
    \mathbf{p}_0 = \mathbf{p}_1,\quad
    \hat{\mathbf{v}} = \frac{\mathbf{p}_2 - \mathbf{p}_1}{\lVert \mathbf{p}_2 - \mathbf{p}_1 \rVert_2}.
\end{equation}
For two lines $\mathcal{L}_1$ and $\mathcal{L}_2$, the smaller intersection angle (hence the absolute value of the dot product) in degrees is
$\theta = \arccos\bigl(|\hat{\mathbf{v}}_1 \cdot \hat{\mathbf{v}}_2|\bigr)\cdot \frac{180}{\pi}$.

\subsection{Landmark Detection as Baseline}\label{sec:method_heatmap_regression}
We chose automatic landmark detection as an intuitive baseline to extract the two points $\mathbf{p}_1$ and $\mathbf{p}_2$ of Eq.~$\eqref{eq:two_pnts_line}$ as the start and end point of the bone structure.
Therefore, we follow the heatmap regression approach \cite{bulat_heatmap_regression_2016}, where a parameterized function $f_k(\phi_k)$, where $\phi_k$ holds all its parameters, maps an input image $\mathbf{I}\in\mathbb{R}^{C\times H \times W}$ where $H$ and $W$ refers to its spatial dimension and $C$ to its number of channels, to two ($K=2$) heatmaps $\mathbf{H}\in\mathbb{R}_+^{K\times H\times W}$.
Here, each pixel $h\propto p(\mathbf{p}_k|\mathbf{I})$ is proportional to the likelihood being the point $\mathbf{p}_k$.
So its coordinate can be obtained by
\begin{equation}\label{eq:heatmap_regression_arg_max}
    \mathbf{p}_k=\underset{h\in H,w\in W}{\arg\max}\ \mathbf{H}(k, h, w).
\end{equation}

\subsection{Bone Pose Estimation with Line Models}\label{sec:point_candidates}
A key limitation of landmark-based bone axis estimation is its high sensitivity to outliers.
If even one of the two defining landmarks is inaccurate, the resulting axis becomes unreliable, which in turn leads to errors in the estimated bone pose.
For this, we extend the parameterized function $f_p(\phi_p)$ to map an image not to two maps holding each the likelihood for a point, but to one single likelihood map $\mathbf{L}\in[0,1]^{H\times W}$ for each pixel $l$ belonging to the axis and hence being a member of $\mathcal{L}$ (cf. Sec.~\ref{sec:line_definition}).
With that, we extract a set of candidates \mbox{$\mathcal{C}=\{l\in \mathbf{L}\mid l> 0.5\}\subseteq \mathcal{L}$} constructing a 2D point cloud.

\subsubsection{Principal Component Analysis with False-Positive Reduction}\label{sec:pca_and_fpr}
Now, with interpreting the set of candidates $\mathcal{C}$ as a 2D point cloud, we can extract the bone structure's axis by applying the principal component analysis (PCA) \cite{pearson_1901_pca} on its zero-mean variant.
Since the eigenvector with the largest eigenvalue captures the largest variance, it can be used as the direction vector $\hat{\mathbf{v}}$ (cf. Sec.~\ref{sec:line_definition}).\par
However, since $\mathcal{C}$ may contain false positives, these outliers can affect the estimated orientation obtained via PCA.
To address this issue, we apply false-positive reduction by exploiting the spatial relationships among the candidate points.
Specifically, we represent the candidates as a binary image analogous to a segmentation mask that highlights the bone structure's axis.
Given this mask, we then can apply well-known mask cleaning operations like morphological operations and connected component analysis (CC).
The first variant (MorphCC) includes opening to remove small isolated false-positives, followed by closing to merge neighboured groups of candidates, with both operations using the same disk structure with radius $r_\text{morph}$ as kernel. 
Finally, we apply a connected component analysis, only keeping the largest one.\par
As an alternative (ThreshCC), omitting morphological operators, we change the component selection from only keeping the largest one by introducing a threshold $\tau$ on the components' area (pixel count).
We define $\tau = \max(\tau_\text{abs}, \tau_\text{rel}\cdot a_\text{largest})$, where $\tau_\text{abs} \in \mathbb{N}$ specifies the minimum component area to be retained, and $\tau_\text{rel} \in [0,1]$ denotes a relative threshold expressed as a fraction of the area of the largest connected component $a_\text{largest}$.

\subsubsection{Random Sample Consensus}
To handle outliers in point candidates $\mathcal{C}$, we can alternatively increase the robustness during fitting a line model by employing Random Sample Consensus (RANSAC) \cite{fischler1981ransac}.
In each iteration, two candidates are randomly sampled from $\mathcal{C}$ to define a line model (cf. Eq.~\eqref{eq:two_pnts_line}).
The residual errors $r$ of all remaining candidates regarding this model are then computed and thresholded using $\tau_\text{R}$ to classify points into inliers and outliers.
After 100 iterations, the model with the highest number of inliers is selected as the final estimate.
To determine $\tau_\text{R}$, we robustly estimate the underlying standard deviation $\sigma_{\mathcal{C}}$ of $\mathcal{C}$.
Since $\mathcal{C}$ may contain outliers, we first perform a loose RANSAC fit using a high threshold $\tau_\text{loose} = 10$.
From this, we obtain the inlier's residuals $\mathcal{R}_\text{loose}=\{r\leq\tau_\text{loose}\mid r\in\texttt{RANSAC}(\mathcal{C})\}$ and estimate $\sigma_\mathcal{C}$ via the median absolute deviation (MAD):
\begin{equation}
    \sigma_\mathcal{C} = 1.4826 \cdot \underset{r\in \mathcal{R}_\text{loose}}{\texttt{median}}(|r-\texttt{median}(\mathcal{R}_\text{loose})|).
\end{equation}
Finally, we obtain $\tau_\text{R}=\sigma_\mathcal{C}\cdot k$ with $k$ as hyperparameter to control the tolerance to outliers.

\subsubsection{Hough Transformation}
Another widely used outlier-robust line model is the Hough transform \cite{duda1972hough}.
To apply it, we first reformulate the line equation in Hesse normal form: $d = x \cos(\varphi) + y \sin(\varphi)$, where $d$ is the distance from the origin to the closest point on the line, and $\varphi$ is the angle between the line and the x-axis.
We then define a function $h(d, \varphi)$ that maps each candidate point into the Hough space, effectively counting “votes” for each $(d, \varphi)$ combination.
The line corresponding to the maximum votes is selected as the final line model:
\begin{equation}
(d^*, \varphi^*) = \underset{d, \varphi}{\arg\max} \ h(d, \varphi).
\end{equation}
Finally, the bone axis direction is obtained as $\hat{\mathbf{v}} =
[\cos(\varphi^*), \sin(\varphi^*)]^\top$.

\section{Experiments}
\subsection{Clinical Task and Datasets}\label{sec:task_and_dataset}
Our experiments cover key paediatric clinical tasks, including fracture fragment angle estimation in both radiographs and ultrasound, and the assessment of developmental dysplasia of the hip using ultrasound.
We use a five-folded cross validation for each dataset, where the first fold was used for hyperparameter tuning and hence excluded from the final evaluation.

\subsubsection{Fracture Fragment Angle Estimation in Radiographs}
We use the public available GRAZPEDWRI-DX dataset \cite{nagy_pediatric_2022} containing over 20k wrist radiographs of 5900 children and adolescents in the AP and lateral view.
We standardized all radiographs by normalizing their laterality to the left side.
The physicians in our team (one radiologist and three paediatric surgeons) extend a subset of 231 radiographs with oriented bounding boxes for the fracture fragments of ulna and radius, resulting in a maximum of four boxes per radiograph.
By averaging the top and bottom corners, we obtain two points representing the axis of a fragment.
Those points were then used to generate the ground truth by drawing a one-pixel line between them (point candidates) or use them directly for the landmark detection.
Subsequently, the fragment axes are used to compute the corresponding angles for fractures of radius and ulna, respectively.
To ensure reproducibility of our results, we include the bounding box annotations in our code repository\textsuperscript{\ref{fn:githublink}}.

\subsubsection{Fracture Fragment Angle Estimation in Ultrasound}
The ultrasound dataset was acquired at UKSH Campus Lübeck between 2019 and 2024 under the Wrist-SAFE protocol \cite{ackermann2019sokrat}, which systematically images the radius and ulna in six longitudinal views each.
It consists of a retrospective part of 43 paediatric patients aged 0 to 16 examined with a GE LOGIC S7, and a prospective part of 221 patients acquired with a Clarius L7HD3 and a GE Venue Go.
Bone and fracture segmentation masks were annotated by medical students under physician supervision.
For this study, a single representative frame was extracted per video.
During preprocessing, masks were filtered via connected component analysis to keep one per fracture label. Samples with fewer than two instances were discarded, and when more existed, the two largest were kept.
Landmarks were defined as the endpoints of each bone mask, corresponding to the farthest left and right x-coordinates of the surface contour.
In total, 147 frames were used for training and evaluation.

\subsubsection{Developmental Dysplasia of the Hip Assessment Using Ultrasound}
The DDH ultrasound dataset was acquired at UKSH Campus Kiel and UMG Göttingen as part of routine neonatal screening using a Clarius L7HD3 linear probe. In total, 133 cases were manually annotated by personnel with medical background under supervision of specialist physicians. For each frame, annotations consist of an ilium segmentation mask and three lines corresponding to the baseline, the bony roofline, and the cartilaginous roofline, from which the alpha and beta angles are derived according to the Graf method. To ensure well-defined line endpoints, start and end points were systematically adjusted using the ilium segmentation mask via a rule-based clipping procedure.

\subsection{Implementation Details}
The functions $f_p$ and $f_k$ are both implemented using a U-Net \cite{Ronneberger2015UNet} (MONAI implementation) with a depth of six stages.
The first stage maps the input image to 32 feature channels, which are subsequently doubled at each stage, resulting in a total of approximately 10 million trainable parameters.
For on-the-fly data augmentation, we apply affine transformations with its parameters randomly sampled for rotation $\mathcal{U}(-15^\circ, 15^\circ)$, translation $\mathcal{U}(-5\%, 5\%)$, and scaling $\mathcal{U}(90\%, 110\%)$.
For the wrist radiographs, we additionally apply random horizontal flipping.
Our code is publicly available on GitHub\textsuperscript{\ref{fn:githublink}}.

\subsubsection{Point Candidates Prediction}
To predict the likelihood of each pixel being a point candidate (cf. Sec.~\ref{sec:point_candidates}), we employ the binary cross-entropy loss using a binary mask as ground truth.
As the goal is to extract the axes of multiple bone structures within a single image, the problem formulation becomes identical to a multilabel segmentation task.
To improve training stability, we increase the one-pixel line width in the ground truth masks by applying morphological dilation with a disk-shaped structuring element of radius 2.
Since the bone structures of interest are present in every image but not always annotated (e.g., healthy bones are not annotated), we restrict backpropagations to samples with non-empty ground truth masks.
This avoids penalizing the model for predicting valid but unlabelled point candidates.

\subsubsection{Landmark Detection}
As described in Sec.~\ref{sec:method_heatmap_regression}, we adopt a heatmap regression approach \cite{bulat_heatmap_regression_2016} for landmark detection, aiming to localize the start and end points of a bone axis.
For each landmark $\mathbf{k} = [h_k, w_k]^\top$, we generate a ground truth heatmap $\mathbf{H}_k \in \mathbb{R}^{H \times W}$ by sampling from a Gaussian distribution:
\begin{equation}\label{eq:heatmap_generation}
    \mathbf{H}_k(h, w)=\alpha_H\cdot\exp\left(-\frac{(h-h_k)^2+(w-w_k)^2}{2\sigma_H^2}\right),
\end{equation}
where $\alpha_H$ and $\sigma_H$ are hyperparameters controlling the amplitude and spatial spread of the Gaussian, respectively.
The model is trained to regress these heatmaps by minimizing the mean squared error loss.
Similar to the point candidate prediction task, backpropagation is performed only for visible (i.e., annotated) landmarks.
To achieve subpixel localization accuracy, we soften the hard $\arg\max$ operation in Eq.~\eqref{eq:heatmap_regression_arg_max} computing a weighted average of the top four highest-response coordinates, where their weights are given by their softmax-normalized heatmap values.

\subsubsection{Hyperparameter Optimization}
For each dataset, we carried out hyperparameter optimization on the first fold of the cross validation split.
We employ two hyperparameter optimization strategies depending on search complexity.
For larger search spaces (>150 trials), particularly involving continuous parameters or combinations thereof, we use the Tree Parzen Estimator (TPE) (Optuna implementation).
This applies to the optimization of $\alpha_H$ and $\sigma_H$ in Eq.~\eqref{eq:heatmap_generation}, as well as the thresholds $\tau_\text{abs}$ and $\tau_\text{rel}$ for connected-component-based false-positive reduction (cf. Sec.~\ref{sec:pca_and_fpr}).
For simpler searches involving single parameters with small discrete ranges, we adopt a grid search strategy.
This includes the morphological radius $r_\text{morph}$ and the RANSAC parameter $k$.
See the supplementary material for the optimized hyperparameters.

\section{Results}
\begin{figure}[t!]
    \begin{subfigure}{\textwidth}
        \centering
        \caption{Statistic reported as $(\mu_{\pm \sigma},\ \mathrm{median})$. Best performance is highlighted in \textbf{bold} (mean) and \underline{underlined} (median). An asterisk ($*$) indicates a statistically significant difference from the best-performing method (Wilcoxon signed-rank test, $p < 0.05$).}
        \begin{tabular*}{\textwidth}{l@{\extracolsep\fill}|ccc}
            & \multicolumn{3}{c}{Task} \\
            Line Model & FracXRay & DDH & FracUS\\\hline
            \protect\tikzblock{tabpurple} PCA & $(6.07_{\pm8.06}, 3.12)$ & $(5.62_{\pm3.29}, 5.08)$ & $(6.86_{\pm9.28}, 3.87)^*$\\
            \protect\tikzblock{taborange} MorphCC + PCA & $\mathbf{(4.10_{\pm3.94}, 3.05)}$ & $(5.41_{\pm3.10}, \underline{4.63})$ & $(6.00_{\pm6.50}, 4.17)$\\
            \protect\tikzblock{tabblue} ThreshCC + PCA & $(4.55_{\pm4.99}, \underline{2.85})$ & $\mathbf{(5.40_{\pm3.15}, 4.92)}$ & $\mathbf{(5.51_{\pm6.78}, \underline{3.25})}$\\
            \protect\tikzblock{tabgreen} Hough & $(4.76_{\pm4.67}, 3.24)^*$ & $(5.97_{\pm3.26}, 5.42)^*$ & $(6.75_{\pm7.21}, 4.10)^*$\\
            \protect\tikzblock{tabbrown} RANSAC & $(4.42_{\pm4.29}, 3.34)$ & $(5.79_{\pm3.33}, 4.85)^*$ & $(6.50_{\pm7.21}, 3.81)^*$\\
            \protect\tikzblock{tabred} Landmark Detection & $(10.02_{\pm8.90}, 7.77)^*$ & $(6.80_{\pm3.66}, 6.21)^*$ & $(6.42_{\pm6.21}, 4.94)^*$\\
        \end{tabular*}
        \label{tab:mae_results}
    \end{subfigure}%
    \vspace{6pt}
    \begin{subfigure}{\textwidth}
        \centering
        \begin{subfigure}[t]{.32\linewidth}
            \includegraphics[width=\linewidth]{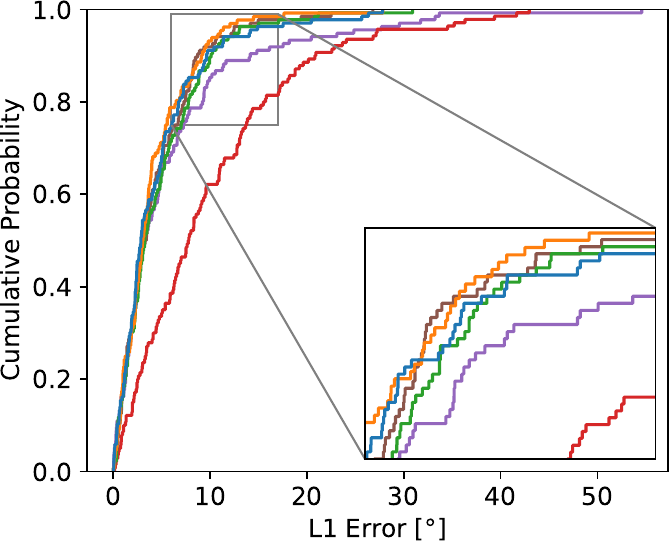}
            \caption{FracXRay}
        \end{subfigure}%
        \hfill
        \begin{subfigure}[t]{.32\linewidth}
            \includegraphics[width=\linewidth]{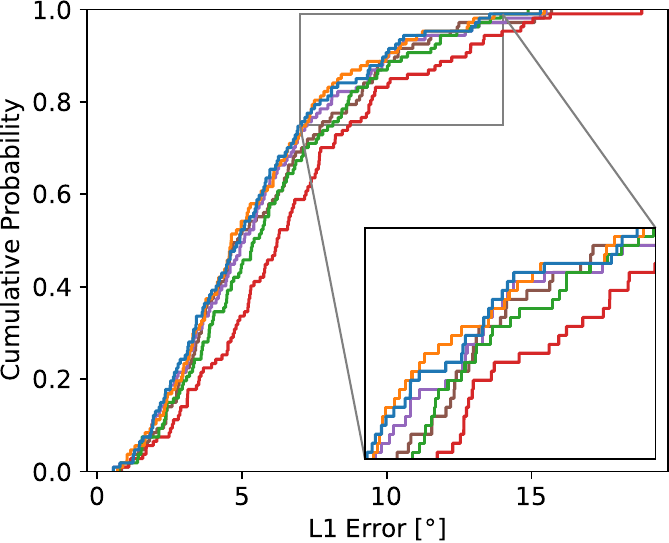}
            \caption{DDH}
        \end{subfigure}%
        \hfill
        \begin{subfigure}[t]{.32\linewidth}
            \includegraphics[width=\linewidth]{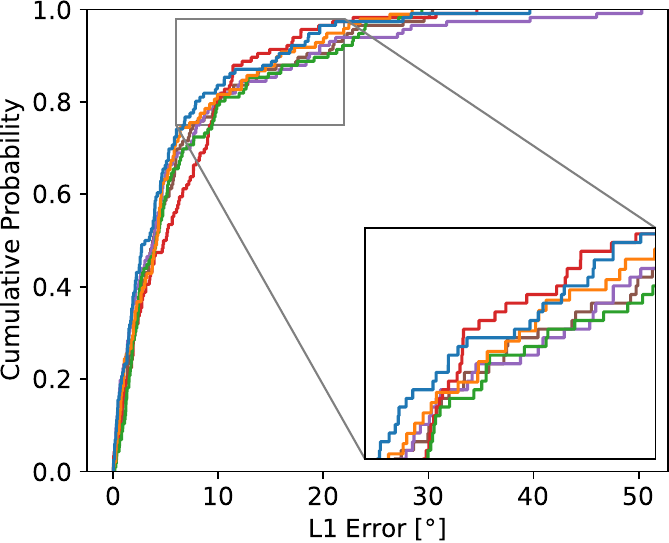}
            \caption{FracUS}
        \end{subfigure}
    \end{subfigure}
    \caption{Statistics of absolute angle estimation error (in degrees) across cross-validation and their empirical cumulative distribution functions.}
    \label{fig:results}
\end{figure}

\begin{figure}
    \centering
    \begin{subfigure}{\textwidth}
        \centering
        \caption*{\textbf{Fracture Fragment Estimation in X-Ray}}
        \begin{subfigure}[t]{.3\linewidth}
            \includegraphics[width=\textwidth]{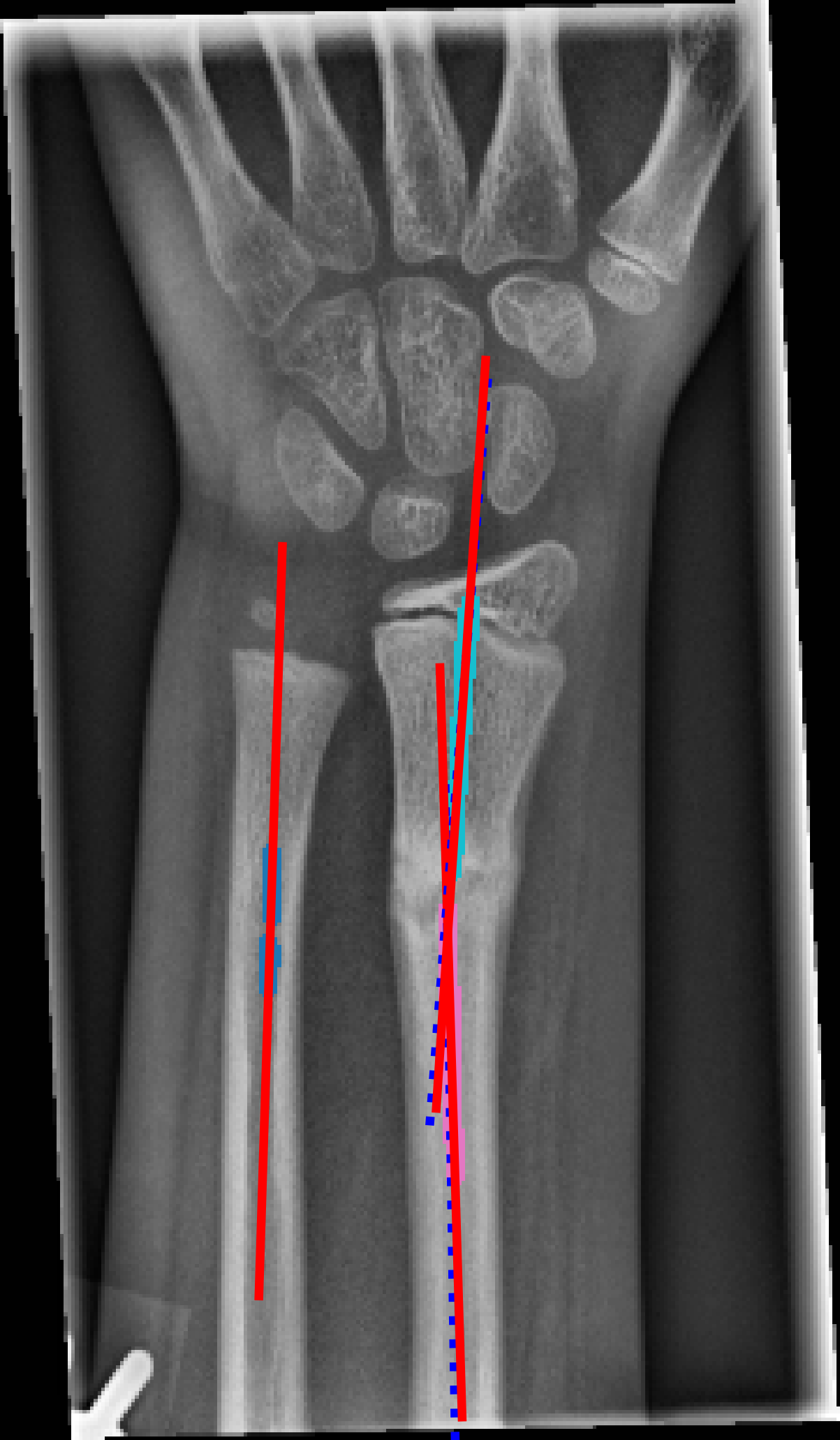}
            \caption{0.13°}
        \end{subfigure}\hfill
        \begin{subfigure}[t]{.3\textwidth}
            \includegraphics[width=\textwidth]{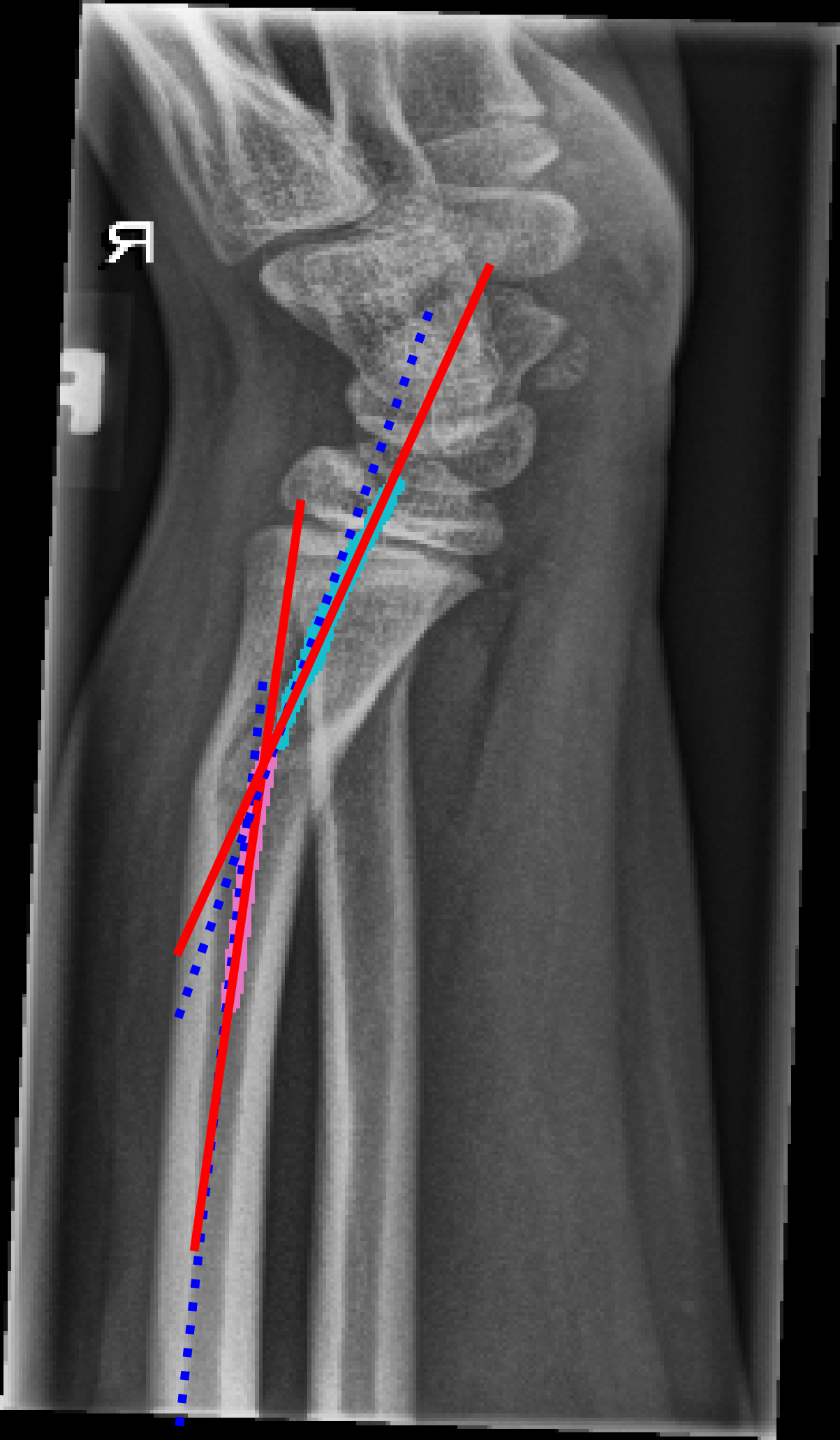}
            \caption{3.05°}
        \end{subfigure}\hfill
        \begin{subfigure}[t]{.3\textwidth}
            \includegraphics[width=\textwidth]{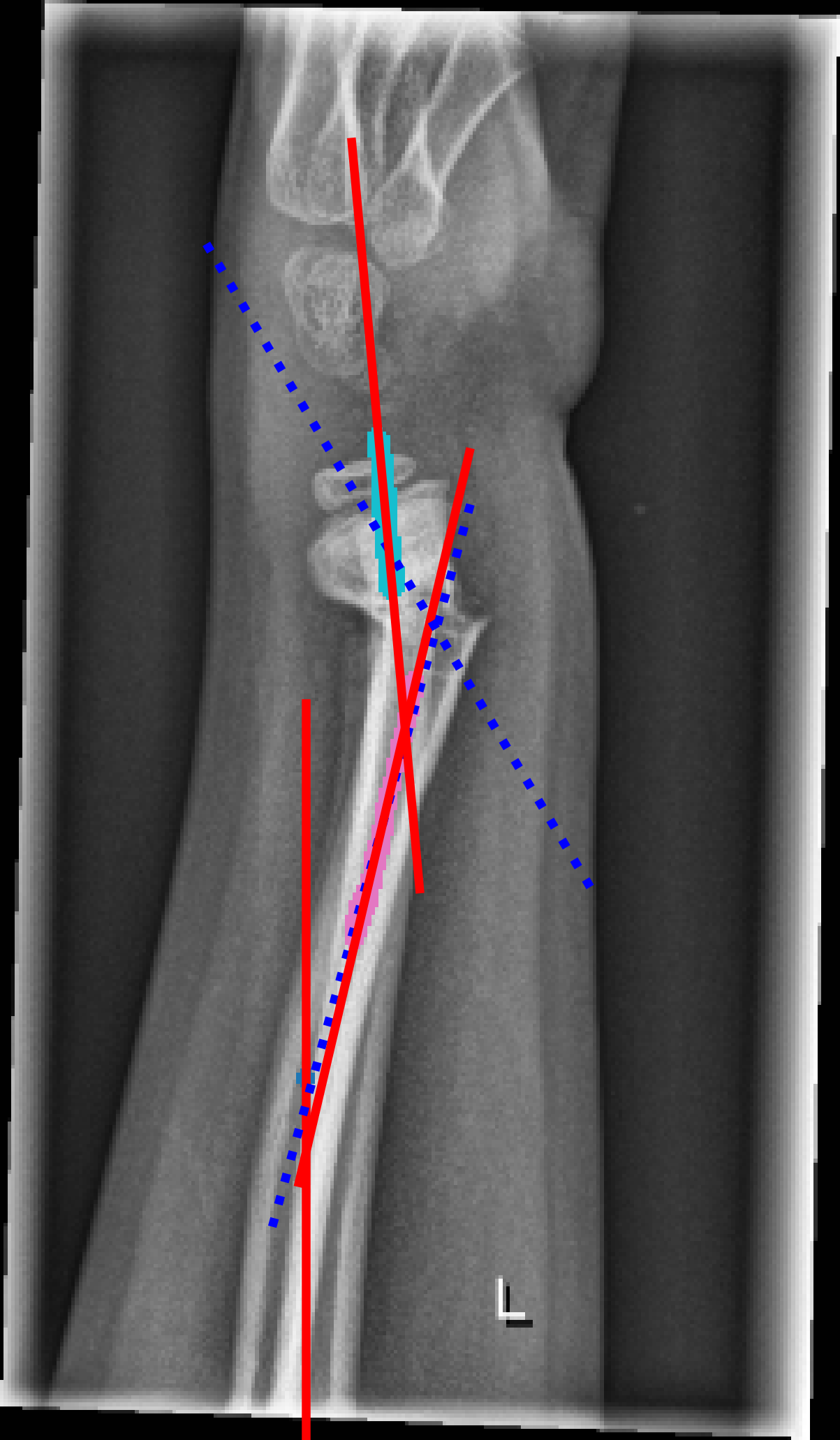}
            \caption{27.84°}
        \end{subfigure}
    \end{subfigure}
    \begin{subfigure}{\textwidth}
        \centering
        \caption*{\textbf{Fracture Fragment Estimation in Ultrasound}}
        \begin{subfigure}[t]{.3\linewidth}
            \includegraphics[width=\textwidth]{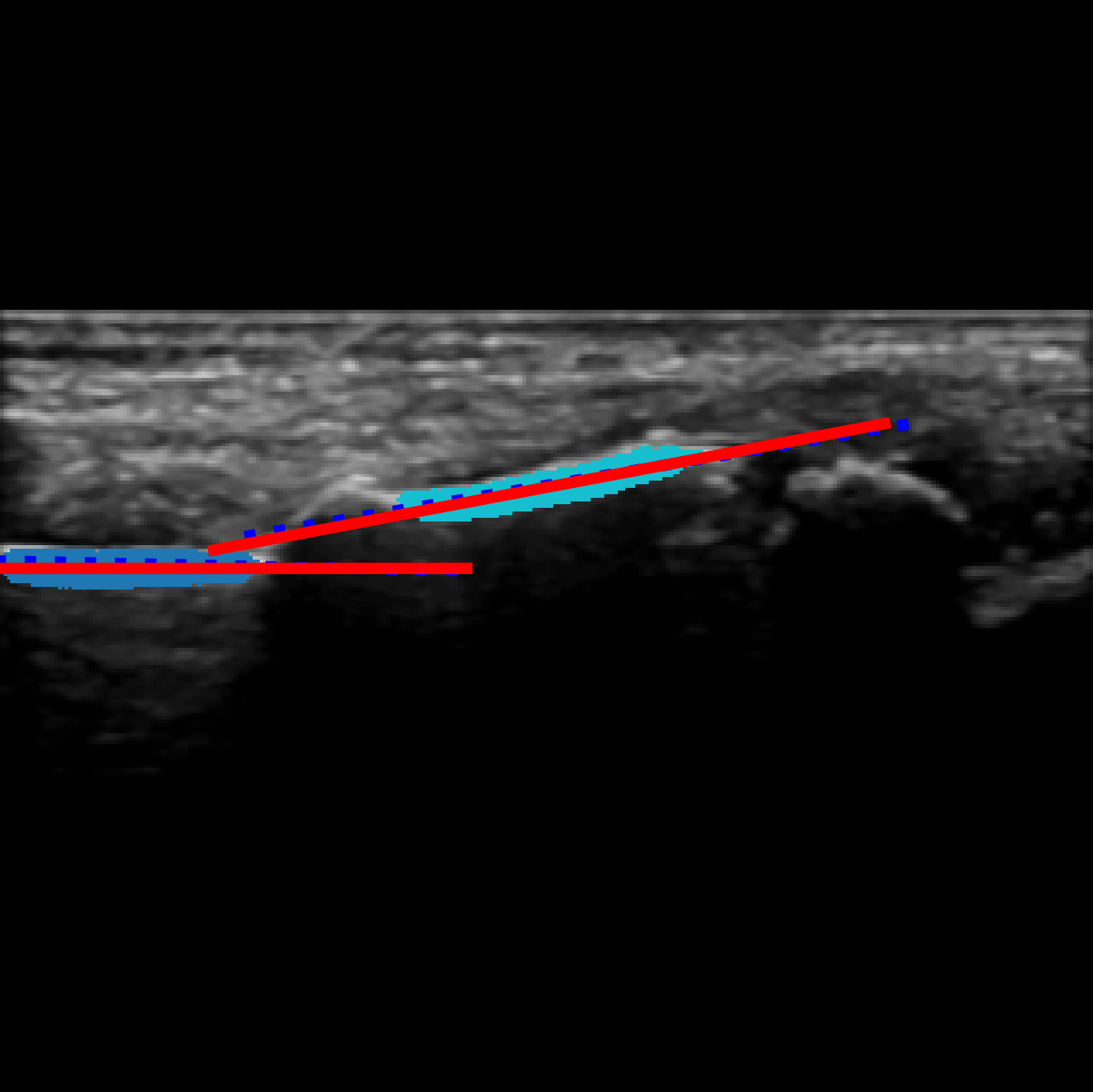}
            \caption{0.04°}
        \end{subfigure}\hfill
        \begin{subfigure}[t]{.3\textwidth}
            \includegraphics[width=\textwidth]{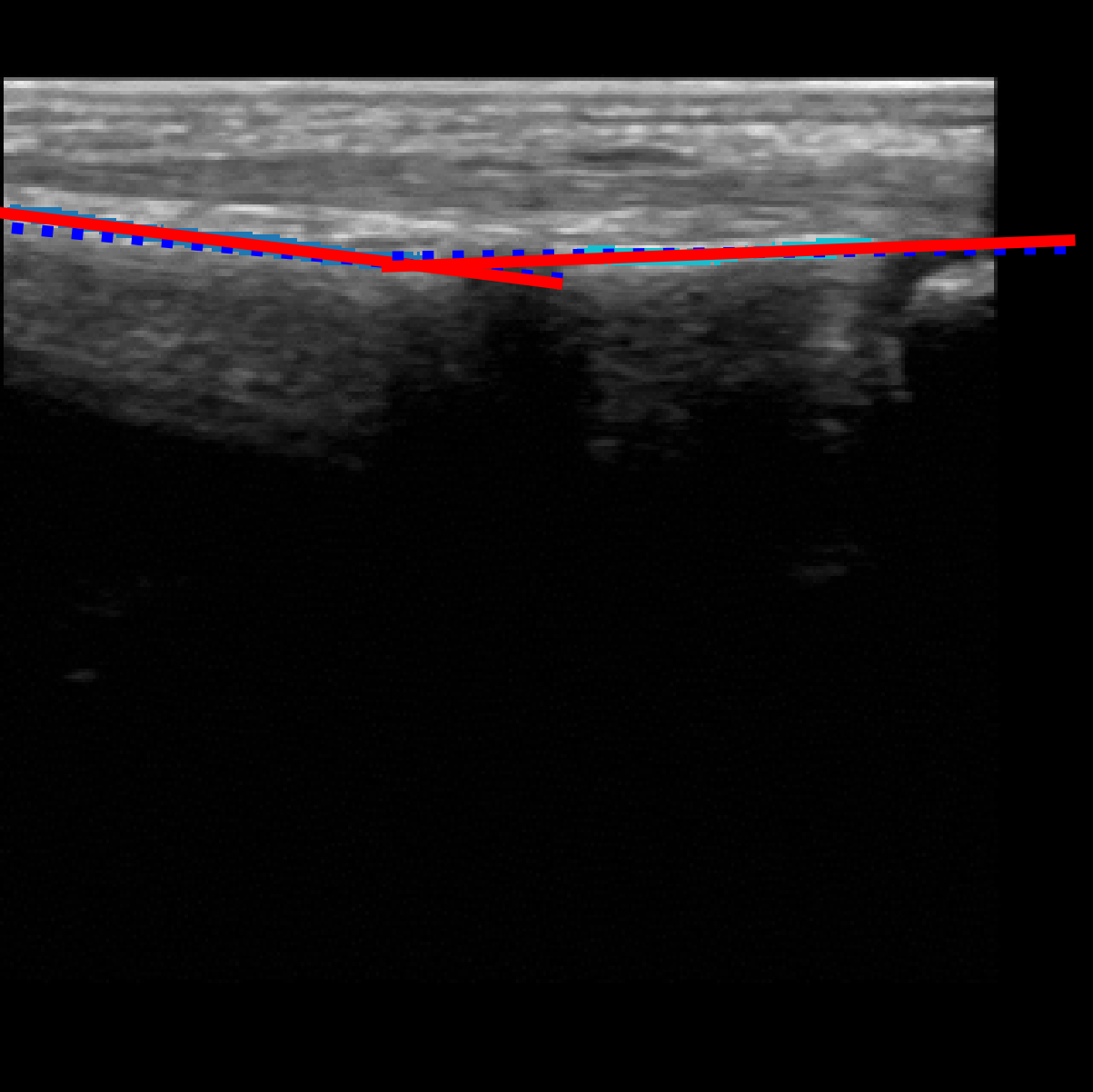}
            \caption{3.25°}
        \end{subfigure}\hfill
        \begin{subfigure}[t]{.3\textwidth}
            \includegraphics[width=\textwidth]{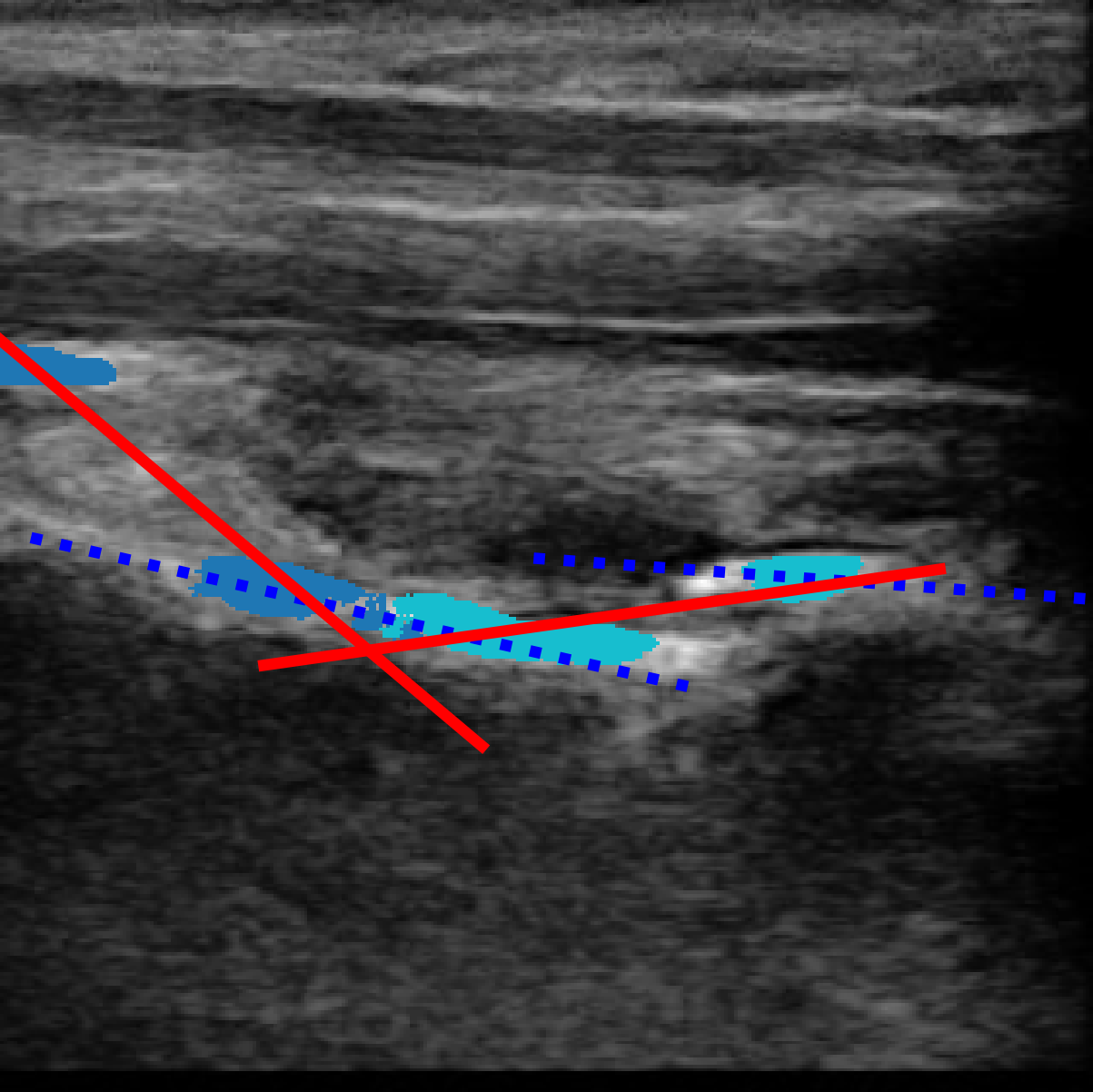}
            \caption{39.65°}
            \label{fig:frac_us_worst}
        \end{subfigure}
    \end{subfigure}
    \begin{subfigure}{\textwidth}
        \centering
        \caption*{\textbf{Developmental Dysplasia of the Hip Assessment Using Ultrasound}}
        \begin{subfigure}[t]{.3\linewidth}
            \includegraphics[width=\textwidth]{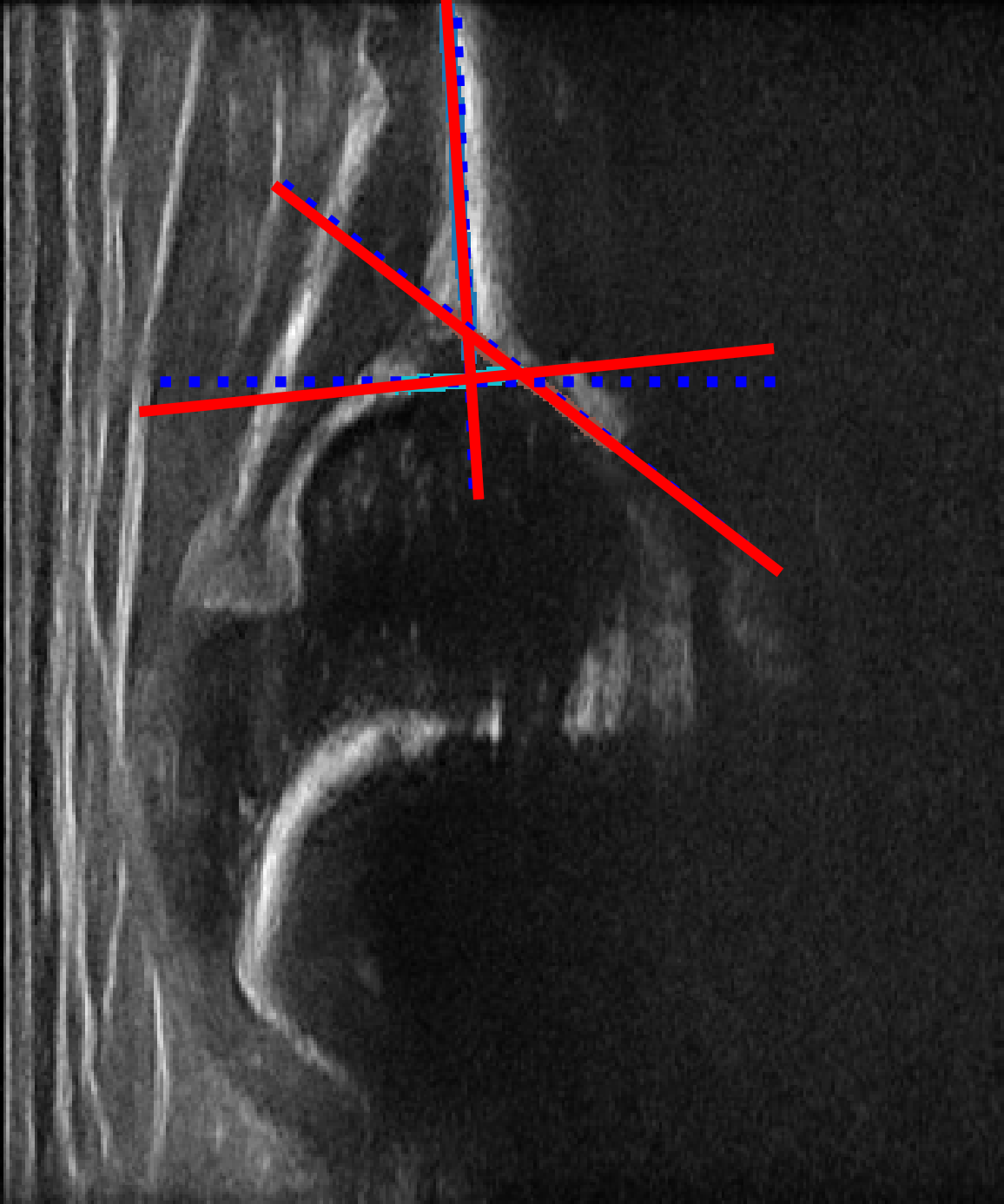}
            \caption{0.56°}
        \end{subfigure}\hfill
        \begin{subfigure}[t]{.3\textwidth}
            \includegraphics[width=\textwidth]{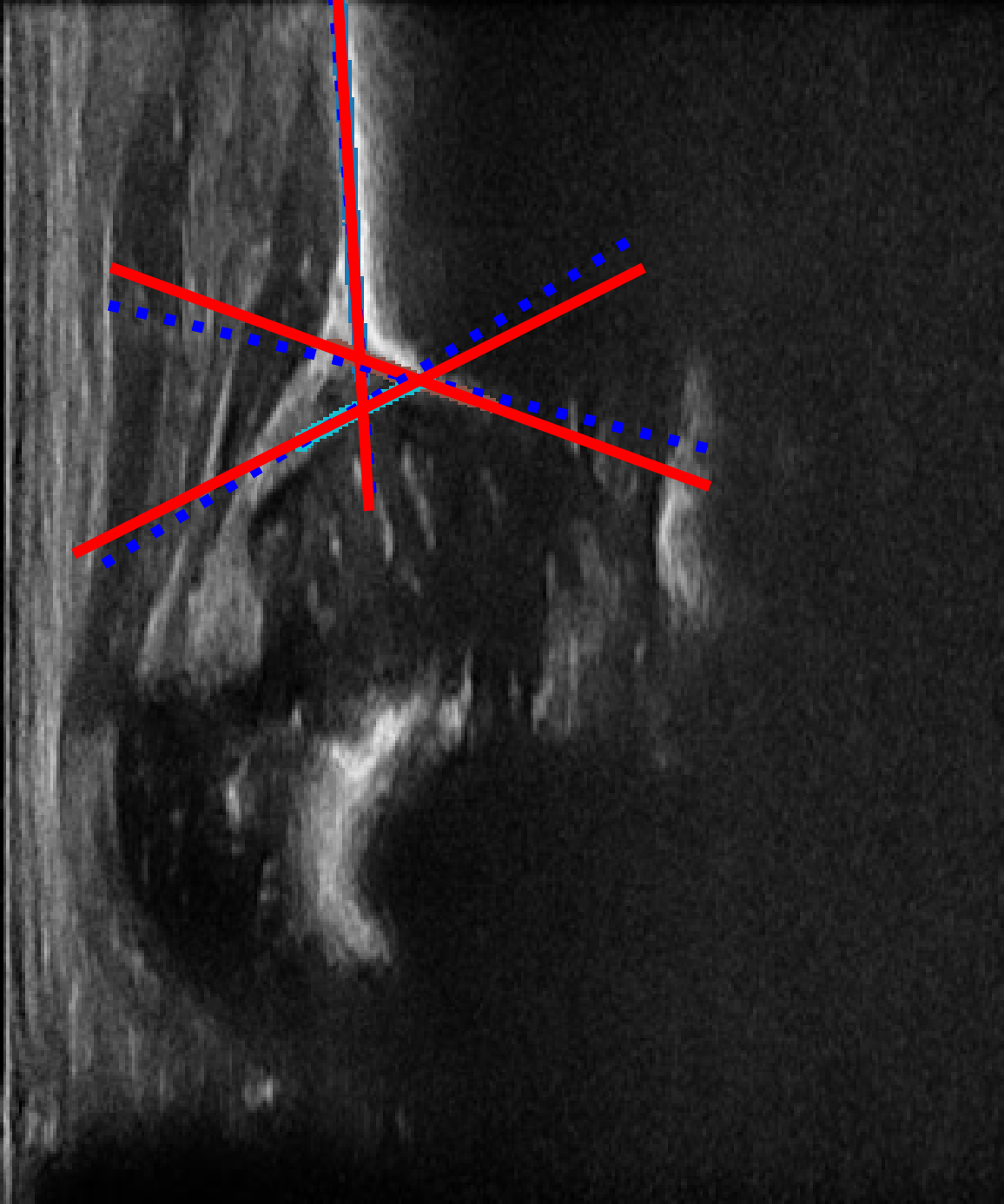}
            \caption{4.92°}
        \end{subfigure}\hfill
        \begin{subfigure}[t]{.3\textwidth}
            \includegraphics[width=\textwidth]{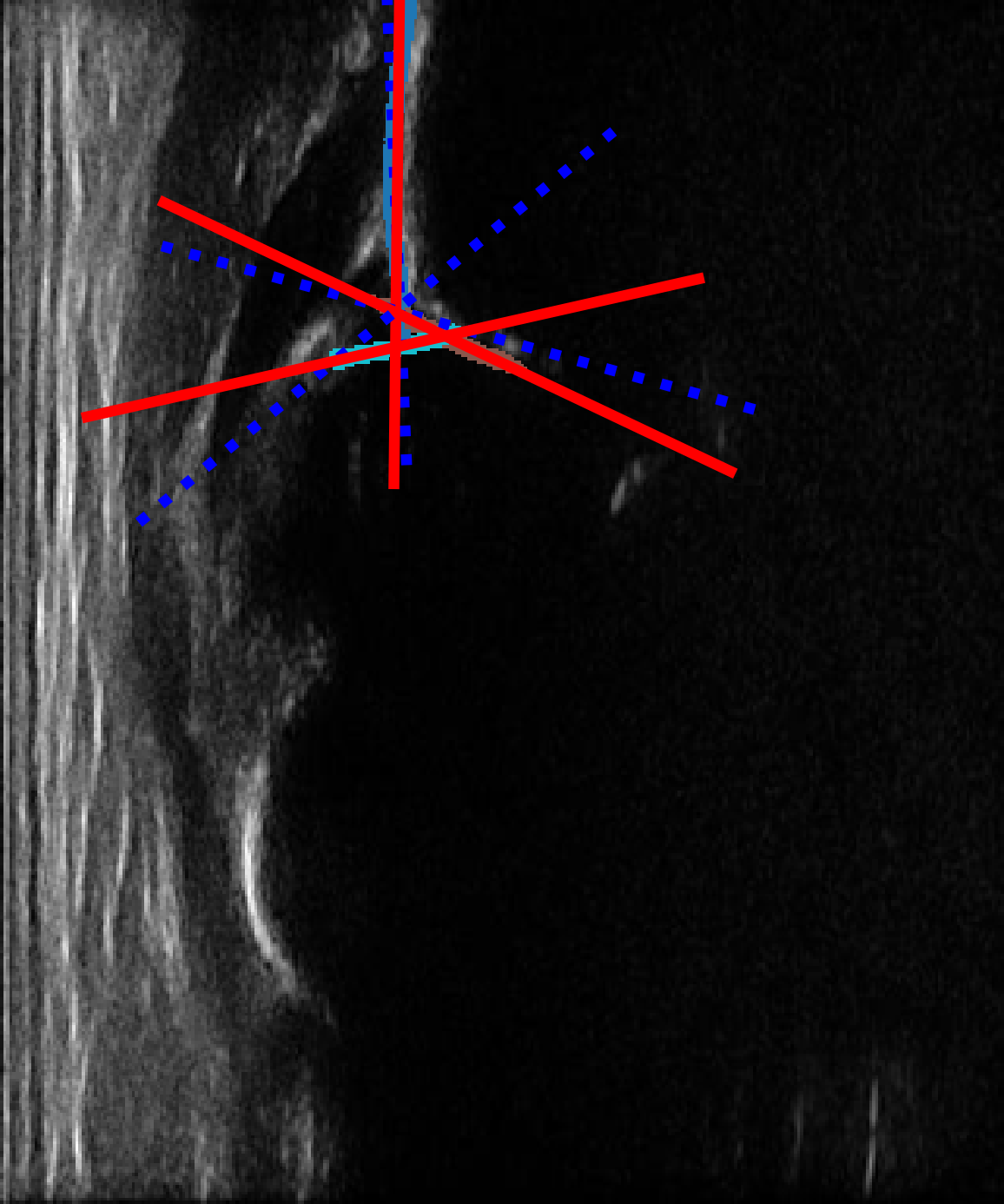}
            \caption{15.30°}
        \end{subfigure}
    \end{subfigure}
    \caption{Qualitative result of the best method (bold in Tab.~\ref{tab:mae_results}) in the three medical angle estimation tasks with the best, median and worst test case (from left to right). Error is given in degree under each case. Coloured masks represent candidate point cloud for axis of different anatomical structures. Red lines represent prediction, blue-dotted the ground truth.}
    \label{fig:qualitative_results}
\end{figure}

Tab.~\ref{tab:mae_results} summarizes the mean absolute error (in degrees) of the estimated angles, averaged over the three paediatric tasks (cf. Sec.~\ref{sec:task_and_dataset}).
Using PCA without false-positive reduction (FPR) already yields strong results, indicating that the U-Net is capable of providing reliable point candidates.
Incorporating FPR further improves performance, with PCA-based line models consistently achieving the best results across all datasets and significantly outperforming most other methods (Wilcoxon signed-rank, $p < 0.05$).
Among the outlier-robust line models, RANSAC consistently performs better than the Hough transform on all three tasks.
In contrast, landmark-based bone axis estimation performs worst on FracXRay and DDH ultrasound.
Compared to the best PCA-based model, errors decrease from $10.02^\circ$ to $4.1^\circ$ on FracXRay and by about $1.4^\circ$ for DDH.
This is also reflected in the eCDFs (Fig.~\ref{fig:results}), where the landmark-based method shows the lowest area under the curve.
For fracture fragment angle estimation in ultrasound (FracUS), it ranks fourth out of six methods.
Further examining the eCDFs, fracture-related tasks show long-tailed distributions, indicating outliers with large errors.
This is reflected in the larger gap between mean and median errors for fractures compared to DDH, highlighting their greater sensitivity to outliers.
While this suggests that fracture fragment axis estimation is generally more challenging, the qualitative examples (Fig.~\ref{fig:qualitative_results}) further illustrate that fracture imaging in both radiographs and ultrasound is acquired in less standardized views, contributing to the increased difficulty.\par

Fig.~\ref{fig:qualitative_results} presents the best, a median, and the worst cases across all three datasets.
For fracture fragment angle assessment in radiographs, the predicted axes generally align well with the actual bone fragment axes.
However, in the worst case, the upper bone fragment is too short for reliable axis estimation, causing misalignment.
Also, the overlap of the radius and ulna in the lateral view, along with their natural curvature, leads to false-positive axis predictions in the lower ulna region.
For bone fragment pose estimation in ultrasound images, both the best and median cases demonstrate that the point candidate suggestions generated by the U-Net are comparable to a segmentation of the bones, enabling precise axis extraction.
In contrast, low image quality and ultrasound artifacts (Fig.~\ref{fig:frac_us_worst}) can split point candidates into two clusters per bone fragment, causing the direction of maximum variance to connect cluster centers rather than align with the true axis.
This highlights a limitation of the ThreshCC variant, where selecting only the largest cluster (as in MorphCC) would have improved accuracy for at least the left fragment.
Qualitative examples for DDH in ultrasound show accurate axis prediction for the iliac bone (upper vertical structure).
However, challenges arise in identifying the capsule (right structure) in the median and worst case, as its deeper position near the hip center reduces contrast and yields less precise point candidates. 

\section{Discussion}
The consistent performance across three paediatric tasks in radiographs and ultrasound highlights the robustness and general applicability of our method combining deep learning-based point candidates with a line model for bone pose estimation.
Moreover, we demonstrate that our approach can be readily extended to handle outliers in the point candidates, either by incorporating established false-positive reduction techniques or by leveraging line-fitting models that are inherently robust to outliers (e.g., RANSAC or the Hough transform) further improving its accuracy.
Importantly, with errors ranging from $4.1^\circ$ to $5.51^\circ$, we operate within the expected clinical observer variability of $5^\circ$ to $8^\circ$ \cite{langensiepen2013scoliosissurvey,gstoettner2007inter}, further supporting its clinical usability.
We attribute the significant superiority of our method over the landmark-based baseline to two key factors:
First, heatmap regression for landmark detection is inherently biased toward image edges, reflecting the inductive bias of convolution.
The elevated error on FracXRay illustrates this limitation, as in this task the start and end points of a fragment axis often lie within homogeneous image regions such as the bone (cf. Fig.~\ref{fig:qualitative_results}, first row).
Second, our representation offers greater robustness.
Even when the set of point candidates includes a proportion of outliers, it can still reliably capture the principal orientation axis of the bone structure.
In contrast, representing it with only two landmarks is inherently more sensitive, as even a slight deviation in a single landmark prediction can lead to a noticeable error in the estimated axis.
Although Fig.~\ref{fig:qualitative_results} illustrates some cases where our method produces misaligned axes, these failures are visually self-evident to the user.
Since the point candidates can also be interpreted as a coarse segmentation of the bone, the reliability of the predicted axis can be intuitively assessed (cf. Fig.~\ref{fig:frac_us_worst}).
However, it should be noted that landmark-based approaches offer similar interpretability, where misaligned landmarks are likewise easy to identify.
Our method was evaluated exclusively on bone structures, which represent the primary clinical application of angle assessment.
Extending this approach to other anatomical structures, however, particularly soft tissue in ultrasound such as the pennation angle of muscle fibres \cite{Strasser2013AssociationBU}, may be more challenging due to lower contrast and less clearly defined boundaries.
A promising direction for future work is the extension to 3D, as all components of our pipeline have been successfully applied in volumetric settings.
This could improve accuracy in applications such as fracture fragment pose estimation by overcoming the perspective distortion inherent in single-plane radiographs \cite{Gu2024Predicting3F}.

\begin{credits}
\subsubsection{\ackname}This research has been funded by the state of Schleswig-Holstein (WTSH, 22023005). The collection and annotation of the ultrasound data has been funded by the German Federal Ministry for Research, Technology, and Space (BMFTR, 13GW0578C and 16SV9252).

\subsubsection{\discintname}
The authors have no competing interests to declare that are
relevant to the content of this article.
\end{credits}
%
%
%
\bibliographystyle{splncs04}
\bibliography{mybibliography}

\newpage

\begin{center}
    \huge \textbf{Supplementary Material: Robust Inter-Bone Pose Estimation in X-Ray and Ultrasound}
\end{center}

\begin{table}[h]
    \centering
    \caption{Results of hyperparameter optimization for each dataset on the first cross-validation fold.}
    \begin{tabular*}{\textwidth}{l@{\extracolsep\fill}|ccc|ccc}
        Method & Parameter & Type & Range & FracXRay & DDH & FracUS \\\hline
        \multirow{2}{*}{Heatmap Regression} & $\alpha_H$ & $\mathbb{N}$ & $[1, 100]$ & 78 & 59 & 32\\
        & $\sigma_H$& $\mathbb{N}$ & $[1, 50]$ & 32 & 11 & 21 \\\hline
        \multirow{3}{*}{False-Positive Reduction} & $r_\text{morph}$ & $\mathbb{N}$ & $[1,8]$ & 1 & 2 & 3\\
        & $\tau_\text{abs}$ & $\mathbb{N}$ & $[1,1000]$ & 1 & 181 & 1\\
        & $\tau_\text{rel}$ & $\mathbb{R}$ & $[0,1]$ & 0.73 & 0.29 & 0.27\\\hline
        RANSAC & $k$ & $\mathbb{R}$ & $[1,8]$ & 1.25 & 1.75 & 1.5
    \end{tabular*}
    \label{tab:hyperparameter}
\end{table}

\end{document}